\documentclass[review]{elsarticle}
\usepackage{amssymb}
\usepackage{amsmath}
\usepackage{graphicx}
\usepackage{lineno}

\begin{document}
	
\ifpreprint
\setcounter{page}{1}
\else
\setcounter{page}{1}
\fi

\begin{frontmatter}
			
\title{VisGraphNet: a complex network interpretation of convolutional neural features}

\author[imecc]{Joao B. Florindo\corref{cor1}} 
\cortext[cor1]{Corresponding author}
\ead{jbflorindo@ime.unicamp.br}

\author[korea]{Young-Sup Lee} 
\ead{ysl@inu.ac.kr}

\author[korea]{Kyungkoo Jun} 
\ead{kjun@inu.ac.kr}

\author[korea]{Gwanggil Jeon} 
\ead{gjeon@inu.c.kr}

\author[ufu]{Marcelo K. Albertini} 
\ead{marcelo.albertini@gmail.com}

\address[imecc]{Institute of Mathematics, Statistics and Scientific Computing - University of Campinas\\
	Rua S\'{e}rgio Buarque de Holanda, 651, Cidade Universit\'{a}ria "Zeferino Vaz" - Distr. Bar\~{a}o Geraldo, CEP 13083-859, Campinas, SP, Brasil}

\address[korea]{
Department of Embedded Systems Engineering, Incheon National University,\\
119 Academy-ro, Yeonsu-gu, Incheon, 22012, Korea	
}

\address[ufu]{
Department of Computer Science, 
Federal university of Uberlandia\\
Av. Joao Naves de Avila 2121, room 1B150, 
Uberlandia, Minas Gerais, Brazil
}

\begin{abstract}
	Here we propose and investigate the use of visibility graphs to model the feature map of a neural network. The model, initially devised for studies on complex networks, is employed here for the classification of texture images. The work is motivated by an alternative viewpoint provided by these graphs over the original data. The performance of the proposed method is verified in the classification of four benchmark databases, namely, KTHTIPS-2b, FMD, UIUC, and UMD and in a practical problem, which is the identification of plant species using scanned images of their leaves. Our method was competitive with other state-of-the-art approaches, confirming the potential of techniques used for data analysis in different contexts to give more meaningful interpretation to the use of neural networks in texture classification.
\end{abstract}


\end{frontmatter}

\section{Introduction}

With a well established theoretical background in physics (statistical physics), mathematics and theoretical computer science (graph theory), complex networks have gathered significant attention from the scientific community since a few seminal works on this topic. Examples of these initial works are \cite{F41,R57,ER59}. Especially after the studies of Watts and Strogatz on small-world networks \cite{WS98}, Barab\'{a}si and Albert on scale-free networks \cite{BAA99} and Girvan and Newman on community structures \cite{GN02}, this model gained more and more attention from applied areas and the complex networks became a widely used technique for the analysis of real world data in several research fields such as in physics \cite{MKKFMGDH19}, medicine \cite{ZMBPHM19}, meteorology \cite{FSSR19}, agriculture \cite{ZWL19}, sociology \cite{BMBL09}, and others.   

More recently, complex networks have also been proposed as a promising tool for the analysis of texture images, more specifically, for the purpose of texture recognition/classification \cite{BCB13}. In the most popular approach for image analysis, the pixels have been interpreted as vertexes of a large-scale graph and measures of similarity among those pixels (difference of gray levels and spatial separation in the image domain, for instance) are modeled as edge weights. Statistical measures like, for example, those associated with the degree distribution, are computed and compose a vector of ``hand-crafted'' descriptors employed as the input of an image recognition system.

Based on this previous success in providing meaningful image descriptors, here we propose the introduction of the complex network modeling in a learning-based framework. More specifically, we propose the use of techniques from the analysis of complex networks to provide a new strategy for pooling image features learned by a neural network composed by convolutional and fully-connected layers. The importance of investigating new strategies for pooling feature maps in neural networks and how this can improve the performance in image recognition tasks has been evidenced by works like \cite{CMKV16,DJVHZTD14}. In this way, we propose the modeling of the fully-connected feature map immediately before the classification outcome as a visibility graph (VG) \cite{LLBLN08}. This is a technique initially proposed for the analysis of time series and converts the feature map into a complex network. From that network, we obtain the image descriptors by computing the degree of each node at different distances. 

The classification accuracy was verified in the recognition of well-established benchmark databases of texture images, namely, KTHTIPS-2b \cite{HCFE04}, FMD \cite{SRA09}, UIUC \cite{LSP05}, and UMD \cite{XJF09}, as well as in an application to the automatic identification of species of Brazilian plants. In all cases, the proposed approach presented promising results and a performance competitive with state-of-the-art methods recently proposed for texture/material recognition.   

\section{Related works}

During the last decades a large number of studies have been presented in the literature using convolutional neural networks (CNN) for image recognition, especially after the impacting work of Krizhevsky et al. in 2012 \cite{KSH12}. That work, in many senses, established baselines for the application of neural networks with a large number of hidden layers (``deep learning'' approach) to the classification of real-world images, achieving performance significantly better than the state-of-the-art at that point in the ILSVCR/ImageNet challenge.

A particular type of image that also benefited from CNN frameworks were texture/material images. Texture recognition tasks can be divided essentially into two types: instance-based and material-based recognition. In instance-based problems we have samples from similar materials varying parameters like viewpoint, illumination, etc. Two typical example of databases in this category are UIUC \cite{LSP05} and UMD \cite{XJF09}. On the other hand, in material-based categorization, the aim is to discriminate among different materials, possibly presenting significant variance within samples. KTHTIPS-2b \cite{HCFE04} and FMD \cite{SRA09} databases illustrate this type of problem.

In instance-based recognition, local ``hand-crafted'' descriptors and ``bag-of-visual-words'' (BoVW) have provided robust results. Local Binary Patterns (LBP) \cite{OPM02} (and its numerous variations), VZ-Joint and MRS4 \cite{VZ09} are examples of such success. More recently an evolution of the BoVW strategy was the Fisher vectors \cite{PD07}, improving the classification accuracy in texture databases and establishing new state-of-the-art records.

For material-based classification, more elaborated techniques were necessary to obtain satisfactory performance. An important ``hand-crafted'' descriptor proposed for this purpose was the dense-SIFT, an adaptation for texture images of the classical Scale-Invariant Feature Transform (SIFT) \cite{L99}, originally developed for object recognition. 

Nevertheless, since the initial tests with ``hand-crafted'' descriptors, their limitations in material classification became clear. Together with the success of deep learning in general purpose image recognition, those were the main motivation for the investigation of CNN-based approaches in texture recognition. However, it was also rapidly verified that the pipeline employed in \cite{KR12} was not the most suitable for texture images. In this case, the use of CNNs as a feature extractor like in DeCAF method \cite{CMKMV14} was more promising. Even more accurate result was achieved by advanced combinations of local descriptors derived from the CNN feature maps with pooling techniques. Examples are the several variations of the combination Fisher vector + CNN (FV-CNN) in \cite{CMKV16}. Finally, we should also mention the CNN architectures adapted for texture images, like T-CNN \cite{AW16}, where another element typically used in hand-crafted descriptors (filter banks) is associated with the CNN framework.

\section{Background}

\subsection{Convolutional neural networks}

Fully connected networks \cite{B95} can be applied to model the task of image classification if we consider each pixel as a neuron in the input layer. However, such a high number of neurons and connections, imply a huge amount of parameters to be optimized, which is computationally costly. Furthermore, the more complex the model (more parameters), the more data we need to reduce the variance of the training set and prevent over-fitting. Moreover, those networks have one dimensional layer, which do not explore local properties of images and the characteristics of each color channel. Convolutional Neural Networks (CNN) are models designed to appropriately deal with images and attenuate all these problems. CNNs allow the network to have neurons organized in higher dimensional layers (3D volumes), exploring the way that pixels are distributed on the image. Some neurons are also allowed to be not connected with others.

After the input data is processed through all the network, the output should be as close as possible to our target outcome. The discrepancy between the expected output and the target is measured by loss functions. The backpropagation algorithm is employed to find the weights that minimize such loss function $Y(x,W)$, which depends both on the input data $x$ and parameters (weights) $W$. This is an iterative process where at each iteration $t$ the weights $w_{ij}$ are updated according to
\begin{equation}
	w_{ij}(t) = w_{ij}(t-1) - \eta \frac{\partial Y}{\partial w_{ij}},
\end{equation}
where $\eta$ is a predefined hyperparameter called \textit{learning rate}.

\subsection{Visibility graph}

Visibility graph (VG) was a tool developed in \cite{LLBLN08} with the original purpose of analyzing time series using techniques from graph theory and complex networks. Generally speaking it maps a sequence of real values to a graph. 

\begin{figure}
	\centering
	\begin{tabular}{cc}
			\includegraphics[width=.5\textwidth]{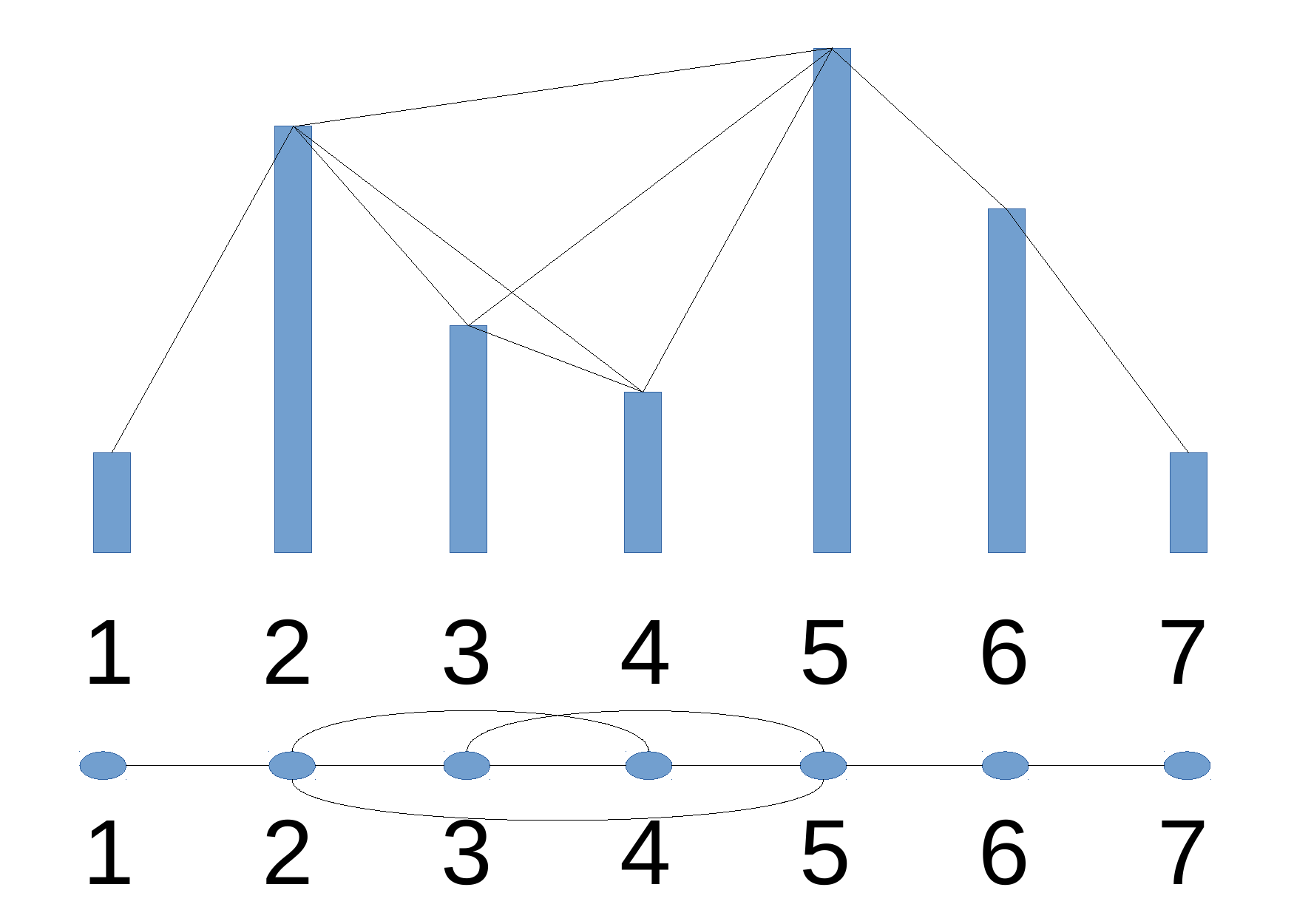} &
			\includegraphics[width=.5\textwidth]{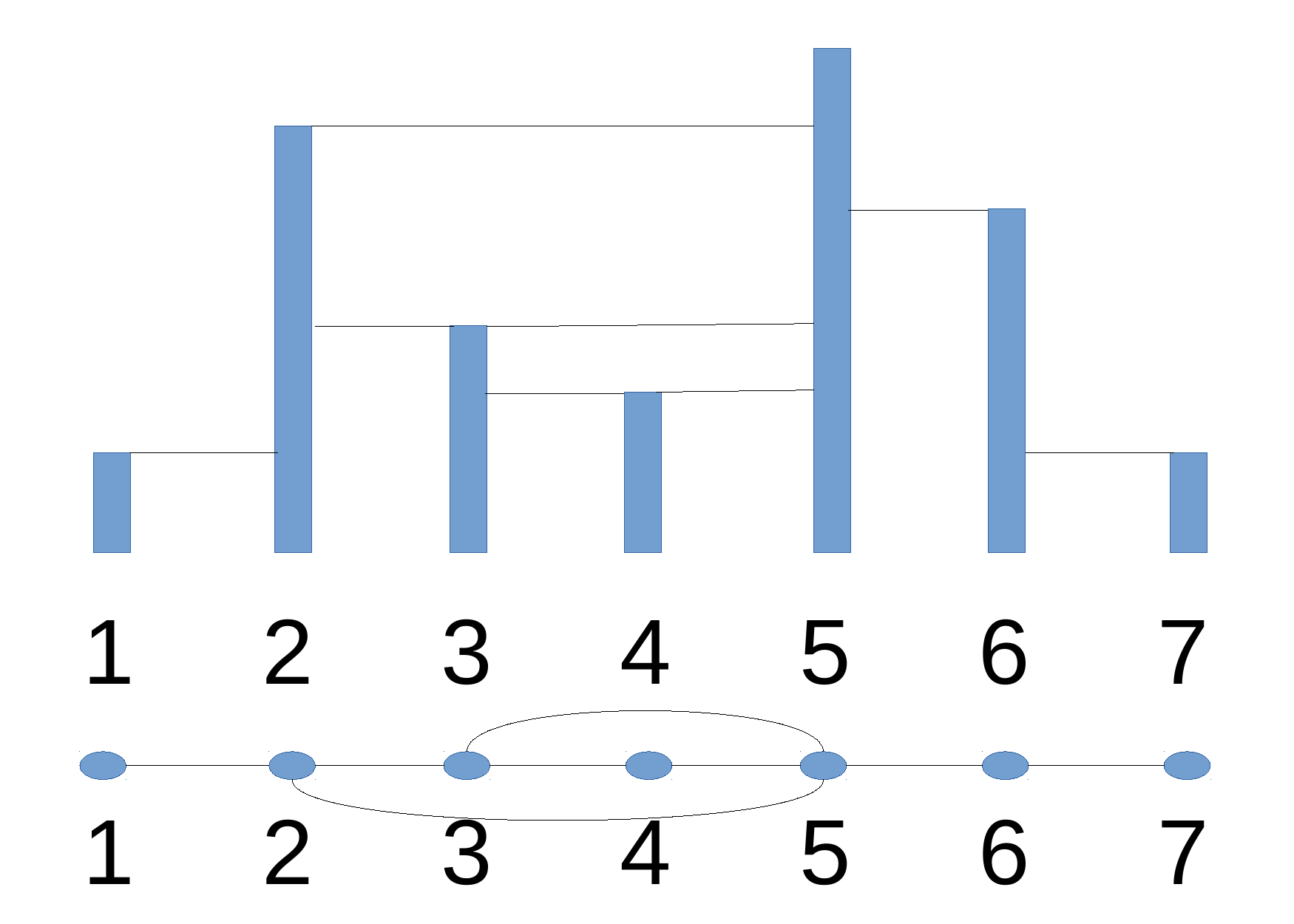}\\
			Natural & Horizontal\\
	\end{tabular}
	\caption{Natural and horizontal visibility graphs. Weighted graphs have the same structure of natural ones, except for having weights associated with each edge.}
	\label{fig:visgraph}
\end{figure}

Figure \ref{fig:visgraph} illustrates the process. Each value is represented by a vertical bar whose height is proportional to the respective value. In terms of the graph, each such bar is associated with a node and two nodes are connected if there is a straight line connecting each vertical bar without intersecting any intermediate bar.

More formally, if we have the value $y_1$ at the position $x_1$ and $y_2$ at $x_2$, we have the corresponding nodes connected if
\begin{equation}
	y_3 < y_2+(y_1-y_2)\frac{x_2-x_3}{x_2-x_1},
\end{equation}
for any $x_3$ between $x_1$ and $x_2$.  

This original version is also known nowadays as Natural Visibility Graphs (NVG). There are two other important variations of VGs: Horizontal Visibility Graphs (HVG) \cite{LLBL09} and Weighted Visibility Graphs (WVG) \cite{SSHJY16}.

In an HVG, we have the points $(x_1,y_1)$ and $(x_2,y_2)$ connected simply if
\begin{equation}
	y_3 < y_1 \mbox{ and } y_3 < y_2,
\end{equation}
for any $x_3$ between $x_1$ and $x_2$. Visually, that means that a horizontal line can connect $(x_1,y_1)$ and $(x_2,y_2)$ without intercepting any other bar. 

Finally, a WVG is pretty similar to an NVG, except that the connection is not expressed by a binary value (``yes'' or ``not''). Instead, it is represented by a real value (weight) that measures the angle of the connecting line. More precisely, the weight $w$ associated to the connection between $(x_1,y_1)$ and $(x_2,y_2)$ is given by
\begin{equation}
	w_{12} = \arctan\frac{y_2-y_1}{x_2-x_1}.
\end{equation}   

\section{Proposed method}

Visibility graphs are known to satisfy several properties that are paramount for pattern recognition tasks. It can be easily demonstrated, for example, that they are invariant under affine transformations, such as horizontal/vertical scaling and translation, rotation, etc. 

There is also a well established relation between the structure of the original data and the topology of the visibility graph. As investigated in \cite{LLBLN08}, periodic data, for example, correspond to regular visibility graphs whose distributions present a finite number of peaks representing the series period. 

On the other hand, a random sequence with values obtained from a uniform distribution results in a random graph, but with an exponential degree distribution (instead of the Poison distribution commonly arising in random graphs). 

Another interesting case is that of fractal-like data, as they often appear in nature modeling. It can be shown (even theoretically in some specific cases) that the degree distribution of the resulting NVG is a power law. The authors in \cite{LLBLN08} illustrate this property with a perfectly self-similar structure. 

This can be demonstrated by constructing a time series starting with the pairs $(x,y)=(0,1)$, $(x,y)=\left( 1,\frac{1}{3} \right)$ and $(x,y)=\left( 2,\frac{1}{3} \right)$. In the next steps they add new pairs placed at random positions but in such a way that, in the step $p$, we have $2^{p+1}$ new pairs, all of them with $y=3^{-(p+1)}$ and with the $x$ values separated by $3^{-p}$. 

For the visibility graph, each pair corresponds to a vertex and for this specific series, the degree can be uniquely identified by the right and left degree of the vertex corresponding to $\left( 2,\frac{1}{3} \right)$. For the right degree at step $n$, here denoted $K_r(n)$, we can use tools from arithmetic functions. The degree function $\mathrm{deg}$ satisfies
\begin{equation}
	\sum_{d|n}\mathrm{deg}(d) = 2^n,
\end{equation} 
where $d|n$ denotes the integer divisors of $n$. In this way, M\"{o}bius inversion formula ensures that at the step $n$
\begin{equation}
	\mathrm{deg}(n) = \sum_{d|n}\mu(d)2^{n/d}.
\end{equation} 
Finally, for the total right degree we should average out over all steps:
\begin{equation}\label{eq:kr}
	K_r(n) = \sum_{k=1}^{n}\frac{1}{n}\sum_{d|n}\mu(d)2^{n/d}.
\end{equation}

For the left degree, the following recursion can be demonstrated by induction:
\begin{equation}\label{eq:kl}
	K_l(n) = 2K_l(n-1) + 1.
\end{equation}

Both expressions (\ref{eq:kr}) and (\ref{eq:kl}) can be summarized into their leading terms, that are exponential relations (power laws):
\begin{equation}
	K_r(n) \approx 2^{4n/5}, \qquad K_l(n) \approx 2^n.
\end{equation}
Power laws are well known in scale-free networks, which also commonly appear in the modeling of real-world problems. 

Another important property of visibility graphs in this case is that the parameter (exponent) of the power law varies according to the Hurst parameter and this is known to be a fundamental feature for distinguishing between different levels of ``fractality''. In this sense, we can say that VGs are also a suggestive tool to infer the fractality of experimental data. 

In summary, VGs are capable of preserving some essential information of the original data (like periodicity, randomness or fractality) at the same time that it offers a new viewpoint such that one can, for example, distinguish data showing simple scale invariance from those presenting small-world characteristics.

\begin{figure}
	\begin{tabular}{ccc}
		\includegraphics[width=.25\textwidth]{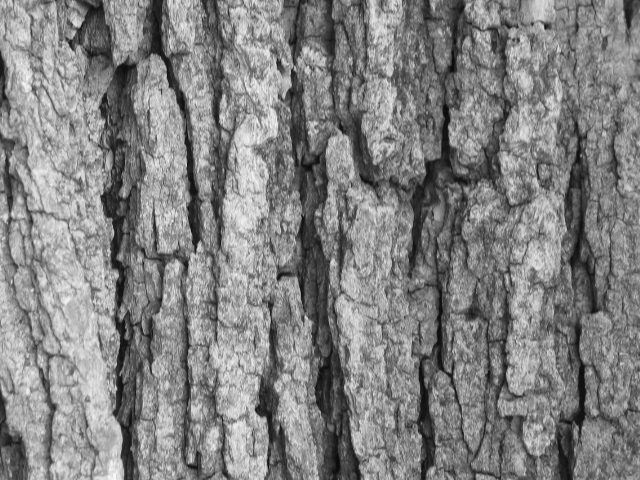} & 
		\includegraphics[width=.25\textwidth]{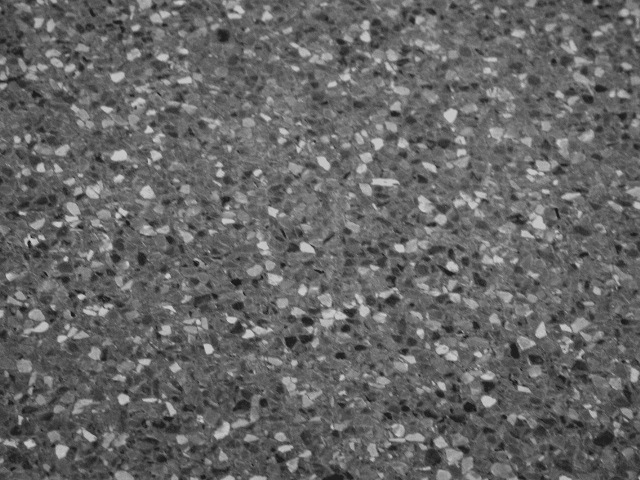} & 
		\includegraphics[width=.25\textwidth]{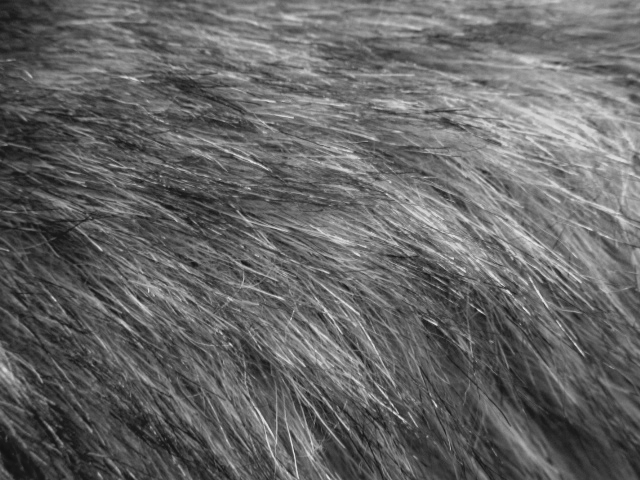}\\
		\includegraphics[width=.3\textwidth]{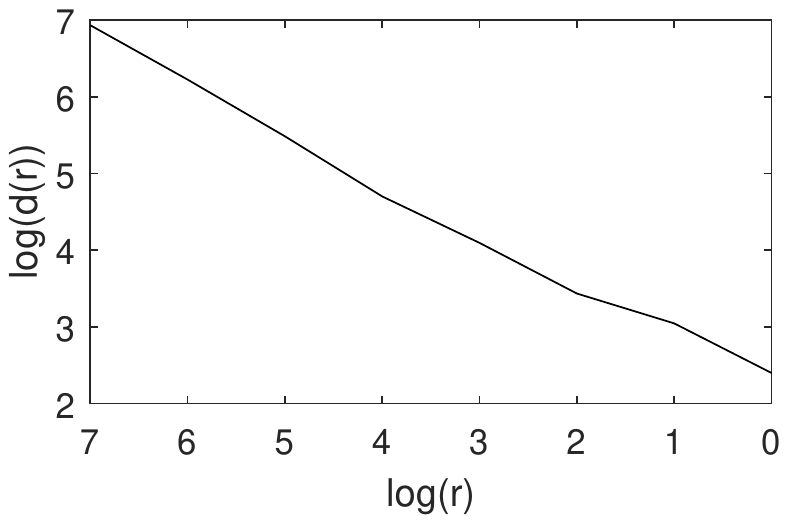} & 
		\includegraphics[width=.3\textwidth]{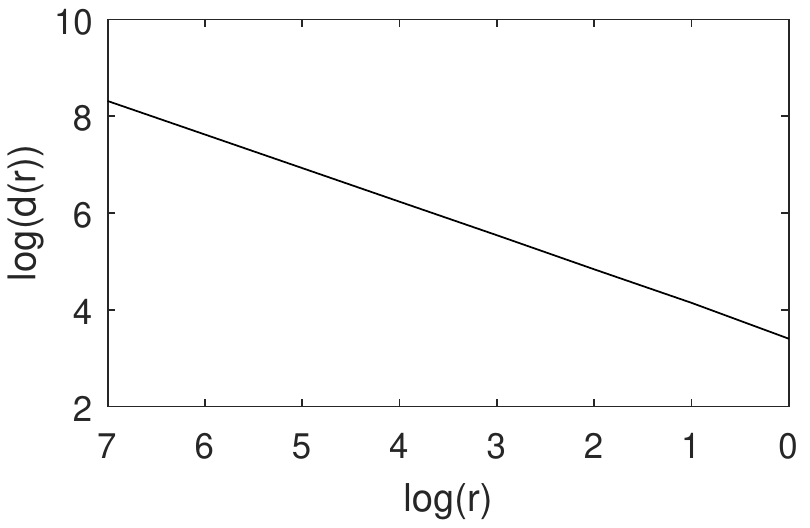} & 
		\includegraphics[width=.3\textwidth]{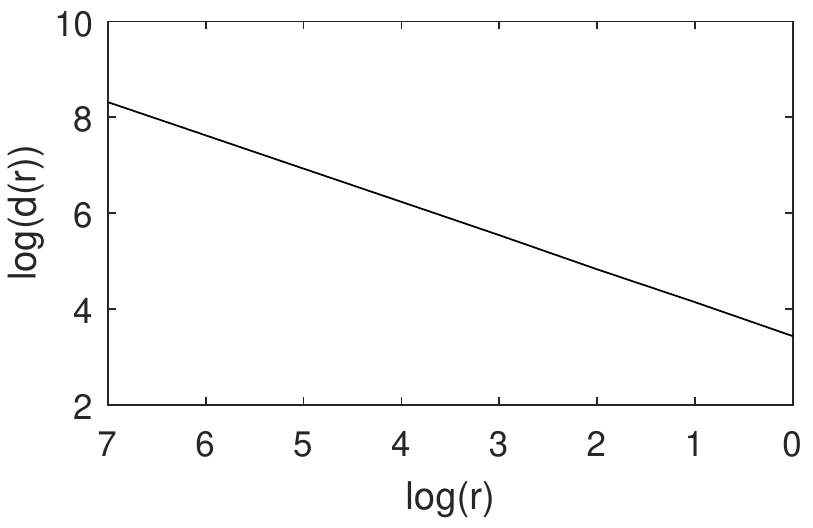}\\		
	\end{tabular}
	\caption{Degree $d$ at different distances $r$ from a particular node in the visibility graph modeling texture features extracted by a convolutional network. The \textit{quasi}-linear form in $\log$-$\log$ scale illustrates the fractality of the images.}
	\label{fig:deg}
\end{figure}

Indeed, properties classically associated to dynamical chaotic systems, like fractal dimension, are well known to provide good description of real-world images \cite{XJF09}. Figure \ref{fig:deg} illustrates how fractal distribution naturally arises in real-world textures such as those in UIUC for example. That figure shows the connectivity degree $d(r)$ of a particular node when values at different distances $r$ in the original data (features extracted by a CNN in this case) are considered. Even though the textures have rather different appearances, in a $\log-\log$ plot the curves in all cases are quite similar to a straight line, confirming the exponential law associated to the degree distribution and attesting its fractality. Nevertheless, while fractal dimension only accounts for the growing rate of a measure along a range of scales, VG graphs are capable of conveying a more complete description of the original data, by including relevant information concerning other possible properties of the features, like randomness, chaoticity and periodicity.

\begin{figure}[!htpb]
	\centering
	\begin{tabular}{cc}
		\includegraphics[width=.2\textwidth]{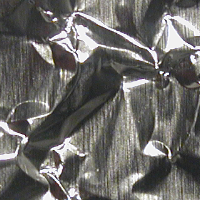} &
		\includegraphics[width=.4\textwidth]{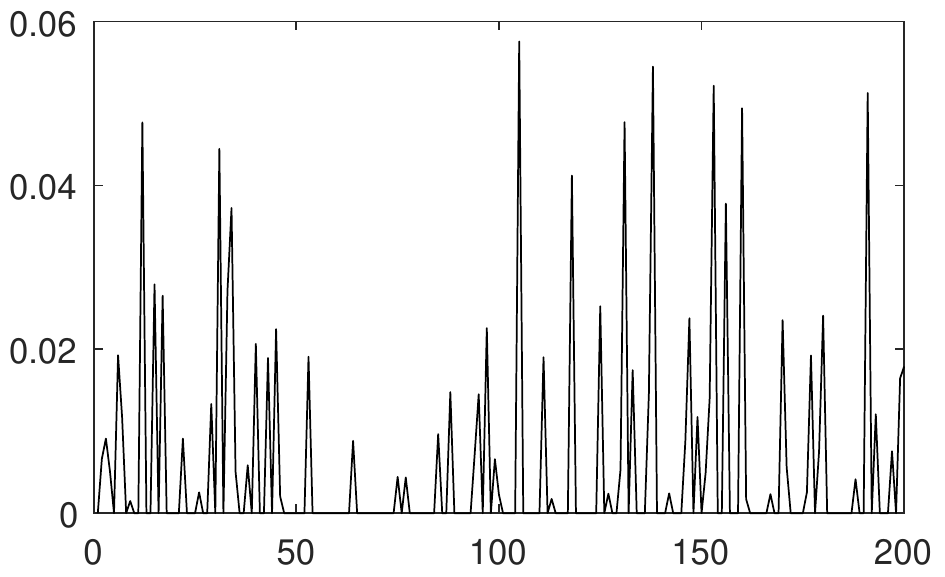}\\
		Image & CNN features\\
		\includegraphics[width=.2\textwidth]{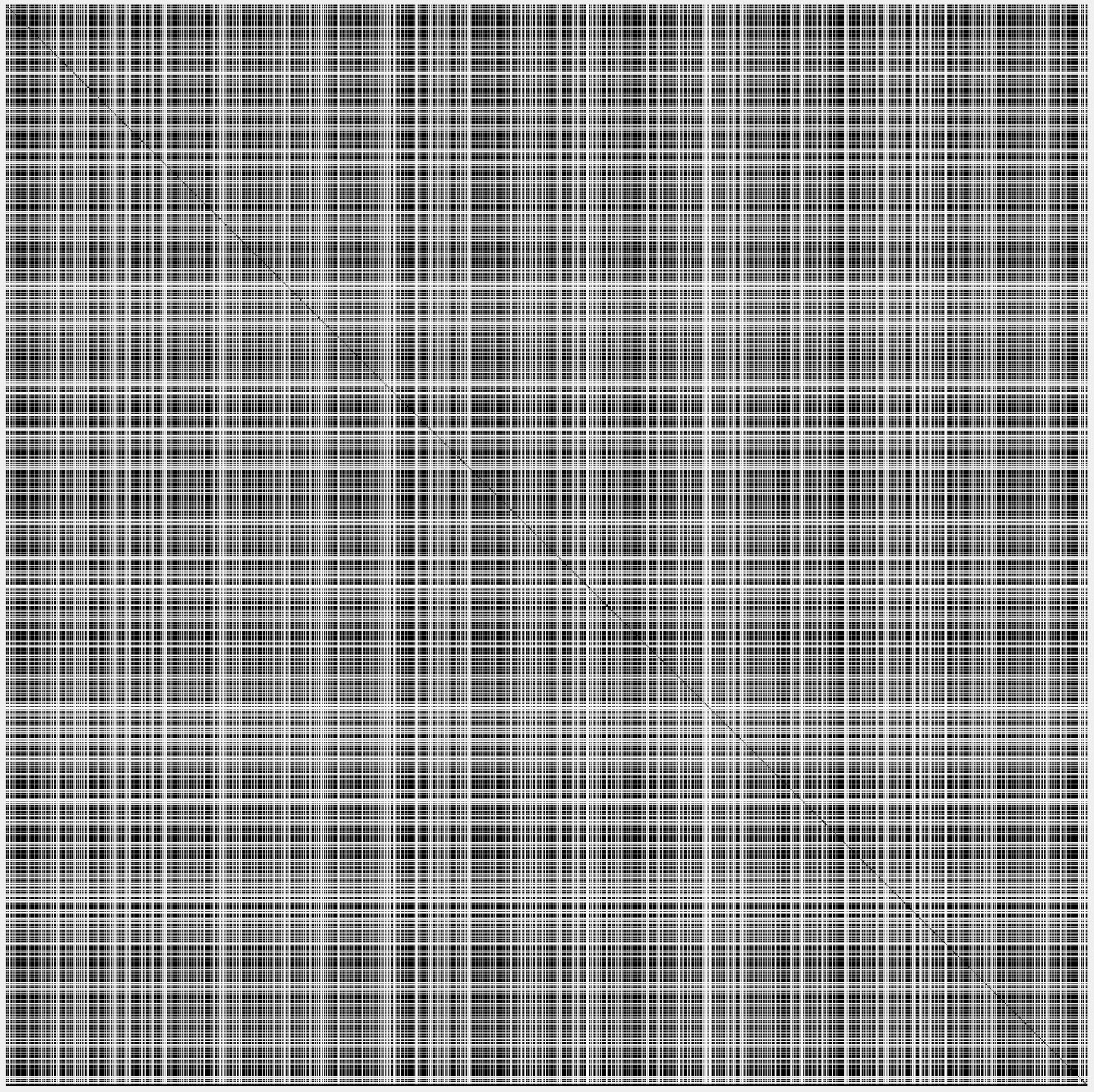} &
		\includegraphics[width=.4\textwidth]{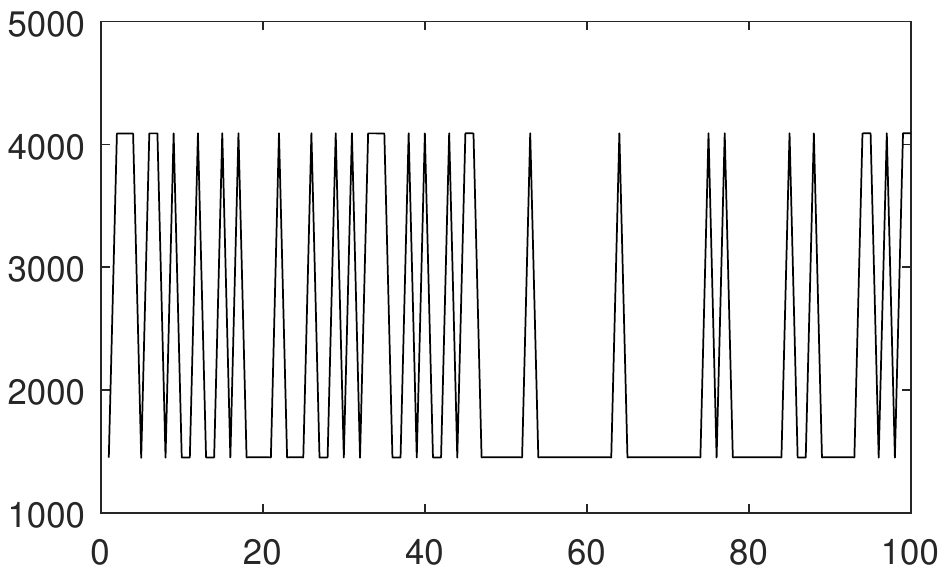}\\
		VG adjacency & Node degree\\
	\end{tabular}
	\caption{VG graph for pooling neural network features from texture images.}
	\label{fig:method}
\end{figure}

This is the motivation for the methodology proposed here, where NVGs, HVGs and WVGs are employed to model the feature map of a neural network at the last fully connected layer. That feature map is used as a descriptor vector for posterior classification with support vector machines in \cite{CMKV16} and provided excellent results. Here the descriptors correspond to the node degrees of the visibility graph. We also consider the degree at different distances, i.e, when rather than counting edges we count paths connecting two vertexes at that particular distance. For weighted VG the meaning of ``degree'' at a particular distance is slightly different, as in this case the product of weights along all paths whose length correspond to that distance are summed up. Figure \ref{fig:method} summarizes the procedure.

For the network architecture we used VGG-VD as that achieved the best performance in most experiments of texture classification in \cite{CMKV16}. We also adopted the transfer learning framework and the network weights were pre-trained on Imagenet database.

For the classification of the proposed descriptors we tested two possibilities: support vector machines (SVM) with a configuration similar to what is employed in \cite{CMKV16}, i.e, linear kernel, $C=1$ and $L^2$ normalization, and linear discriminant analysis (LDA) \cite{K88}.

\section{Experiments}

The performance of the proposed descriptors was evaluated on four benchmark databases widely used in the recent literature of texture/material classification methods. We also applied the same method to a practical problem, namely, the identification of species of Brazilian plants using scanned images of leaves.  

The first benchmark data set is KTHTIPS-2b \cite{HCFE04}, a database comprising 4752 images equally divided into 11 material categories. An important characteristic of this data is its focus on the material represented in the image rather than on the instance of the photographed object. In each material class the images can still be divided into 4 samples. Each sample follows a particular scheme of scale, pose and illumination. The validation protocol is the most typically employed in the literature, where 1 sample is used for training and the remaining 3 samples are used for testing. The accuracy (percentage of images correctly classified) is obtained by averaging out the results for the 4 possible combinations of training/testing.

The second database is Flickr Material Database (FMD) \cite{SRA09}. These images were manually collected from flickr.com aiming at representing materials more commonly found in our daily lives and at the same time capturing variance in illumination, color, composition, etc. The database contains 1000 images equally divided into 10 categories. Again we use the protocol typically followed in the literature, with 14 random training/testing splits. Each split contains 50 images per class for training and the remaining 50 images for testing.

The third database is UIUC \cite{LSP05}, containing 1000 images evenly divided into 25 texture categories. The images were collected under non-controlled conditions and contain variation in albedo, perspective, illumination and scale. For the validation split we randomly select 20 images of each texture class for training and the remaining 20 images for testing. This is repeated 10 times to provide the average accuracy.

The fourth data set is UMD \cite{XJF09}. This is in many respects similar to UIUC, containing the same number of classes and images per class, and similar challenges like uncontrolled conditions of acquisition and variation in viewpoint and scale. The most significant difference is that in this case the images have higher resolutions, with $1280\times 960$ pixels.

\section{Results and Discussion}

As a first test we compared the performance of two well known classifiers, i.e., LDA and SVM, on the proposed descriptors. The features employed in this case are given by the degree values of natural VG at distance 1. The descriptors are also normalized to fit within $0$ and $1$. Table \ref{tab:classif} lists the average accuracy (percentage of images correctly classified) among the training/testing splits achieved in each case. LDA performed better in all databases and as a consequence it was the classifier employed for the remaining tests.
\begin{table}[!htpb]
	\centering
	\caption{Classification accuracy (\%) for the proposed descriptors using SVM and LDA classifiers.}
	\label{tab:classif}
	\begin{tabular}{ccc}
		\hline
		Database & SVM & LDA\\
		\hline
		KTHTIPS-2b & 71.1 & 75.7\\
		FMD & 73.5 & 77.3\\
		UIUC & 97.2 & 97.6\\
		UMD & 97.8 & 98.1\\
		1200Tex & 84.5 & 87.4\\		
%		KTHTIPS-2b & 71.1$\pm$5.6 & 75.6$\pm$1.8\\
%		FMD & 73.5$\pm$2.2 & 76.9$\pm$1.8\\
%		UIUC & 97.2$\pm$1.5 & 97.6$\pm$0.7\\
%		UMD & 97.8$\pm$1.1 & 98.1$\pm$0.8\\
%		1200Tex & 84.5$\pm$1.8 & 86.7$\pm$1.4\\
		\hline		
	\end{tabular}
\end{table}

Table \ref{tab:param1} shows the percentage of images correctly classified when using the three types of visibility graphs investigated here and at distances 1, 2 and 3. A few points should be highlighted here. The fist one is that in most cases horizontal graphs provided higher accuracies than natural and weighted VGs. This is justified by the simplification accomplished by HVG in comparison with NVG or WVG, as a few edges are removed. As a consequence, the resulting descriptors correspond to a more global perspective of the original data, being thus less prone to over-fitting. The results for weighted graphs demonstrated that the angle of the visibility connection is not sufficiently relevant as a representation of the CNN codes. This is potentially explained by the introduction of an artificial periodic behavior by the angle measure, which is not present in the original data. Other types of weights for the visibility connection might be investigated in the future to circumvent this issue. It is also interesting to observe that there is no performance gain with the addition of larger distances, in fact, the larger the distance the lower was the accuracy. This was also the rationale for discarding tests with even larger distances, as they imply more computation burden associated with no improvement in the accuracy.
\begin{table}[!htpb]
	\centering
	\caption{Classification accuracy (\%) for the VisGraphNet descriptors, type 'natural' (N), 'horizontal' (H) and 'weighted' (W) at distances 1 (D1), 2 (D2) and 3 (D3).}
	\label{tab:param1}
	\scalebox{0.9}{
	\begin{tabular}{cccccccccc}
		\hline
		Database & ND1 & ND2 & ND3 & HD1 & HD2 & HD3 & WD1 & WD2 & WD3\\
		\hline
		KTHTIPS-2b & 75.7 & 75.3 & 74.7  & \textbf{77.0}  & 76.6  & 76.8& 75.2  & 72.3  & 70.6\\
		FMD & 77.3 & 76.3 & 75.0  & \textbf{77.5}  & 77.5  & 76.8 & 77.2$\pm$1.8 & 71.4 & 67.1\\
		UIUC & 97.6 & 97.2 & 96.2  & \textbf{97.8} & 97.7  & 97.7 & 97.7 & 95.9 & 93.4\\
		UMD & 98.1 & 97.9 & 98.0 & \textbf{98.5} & 98.3 & 98.1 & 98.0 & 97.5 & 96.5\\		
		1200Tex & \textbf{87.4} & 86.0 & 85.0 & 87.3 & 86.9 & 86.1 & 87.0 & 86.7 & 85.3\\		
%		KTHTIPS-2b & 75.7$\pm$1.9 & 75.3$\pm$1.9 & 74.7$\pm$1.9  & \textbf{77.0$\pm$1.7}  & 76.6$\pm$1.8  & 76.8$\pm$1.3\\
%		FMD & 77.3$\pm$1.5 & 76.3$\pm$1.7 & 75.0$\pm$2.0  & \textbf{77.5$\pm$1.6}  & 77.5$\pm$1.8  & 76.8$\pm$2.7\\
%		UIUC & 97.6$\pm$0.8 & 97.2$\pm$1.1 & 96.2$\pm$0.7  & \textbf{97.8$\pm$0.9} & 97.7$\pm$0.6  & 97.7$\pm$0.7\\
%		UMD & 98.1$\pm$0.9 & 97.9$\pm$0.9 & 98.0$\pm$0.9 & \textbf{98.5$\pm$0.7} & 98.3$\pm$0.7 & 98.1$\pm$0.6\\		
%		1200Tex & \textbf{87.4$\pm$1.2} & 86.0$\pm$1.2 & 85.0$\pm$1.3 & 87.3$\pm$1.1 & 86.9$\pm$1.3 & 86.1$\pm$1.7\\
		\hline		
	\end{tabular}
	}
%	\begin{tabular}{cccc}
%		\hline
%		Database & WD1 & WD2 & WD3\\
%		\hline
%		KTHTIPS-2b & 75.2$\pm$1.6  & 72.3$\pm$1.4  & 70.6$\pm$1.3\\
%		FMD & 77.2$\pm$1.8 & 71.4$\pm$2.3 & 67.1$\pm$2.7\\
%		UIUC & 97.7$\pm$0.6 & 95.9$\pm$0.9 & 93.4$\pm$1.4\\
%		UMD & 98.0$\pm$0.9 & 97.5$\pm$1.2 & 96.5$\pm$1.1\\		
%		1200Tex & 87.0$\pm$1.5 & 86.7$\pm$1.4 & 85.3$\pm$1.5\\
%		\hline		
%	\end{tabular}
\end{table}

Given that lower distances yielded better results, Table \ref{tab:param2} shows the classification accuracies when distances 1 and 2 are combined for each VG type. Horizontal graph repeated the great performance in Table \ref{tab:param1}, however now we also have good results achieved by weighted VGs. This can be explained by the fact that the distance degree in this case conveys more information than a binary adjacency matrix. Now the weights along any path with that particular length are accounted and the way how information for D1 and D2 complement each other is more evident.
\begin{table}[!htpb]
	\centering
	\caption{Classification accuracy (\%) for the VisGraphNet descriptors combined at distances 1 and 2.}
	\label{tab:param2}
	\begin{tabular}{cccc}
		\hline
		Database & ND1+ND2 & HD1+HD2 & WD1+WD2\\
		\hline
		KTHTIPS-2b & 75.8 & \textbf{77.2} & 75.3\\
		FMD & 75.9 & \textbf{77.5} & 77.4\\
		UIUC & 96.9 & 97.8 & \textbf{97.9}\\
		UMD & 97.9 & \textbf{98.4} & 98.3\\		
		1200Tex & 85.9 & 86.9 & \textbf{86.9}\\		
%		KTHTIPS-2b & 75.8$\pm$2.3 & \textbf{77.2$\pm$1.4} & 75.3$\pm$1.6\\
%		FMD & 75.9$\pm$2.3 & \textbf{77.5$\pm$1.8} & 77.4$\pm$1.8\\
%		UIUC & 96.9$\pm$0.9 & 97.8$\pm$0.9 & \textbf{97.9$\pm$0.5}\\
%		UMD & 97.9$\pm$0.8 & \textbf{98.4$\pm$0.8} & 98.3$\pm$0.8\\		
%		1200Tex & 85.9$\pm$1.1 & 86.9$\pm$1.7 & \textbf{86.9$\pm$1.1}\\
		\hline		
	\end{tabular}
\end{table} 

Table \ref{tab:param3} shows two other combination of VG descriptors: all graphs at distance 1 and horizontal VG at distance 1 combined with degree sequence (DS), i.e., the list of degree values sorted in increasing order. The combination with DS gave some boost to the horizontal descriptors. Such extra discrimination can be explained by an intrinsic property of degree sequences according to which two graphs with different DS cannot be isomorphic, i.e., they possess real differences in their topologies. 
\begin{table}[!htpb]
	\centering
	\caption{Classification accuracy (\%) for the VisGraphNet descriptors combining types N, H and W at distance 1 and combining H at distance 1 with degree sequence (DS).}
	\label{tab:param3}
	\begin{tabular}{ccc}
		\hline
		Database & N1+H1+W1 & H1+DS\\
		\hline
		KTHTIPS-2b & 75.7 & \textbf{77.5}\\
		FMD & \textbf{77.3} & 77.3\\
		UIUC & 97.5 & \textbf{98.0}\\
		UMD & 98.1 & \textbf{98.4}\\		
		1200Tex & \textbf{87.8} & 87.3\\		
%		KTHTIPS-2b & 75.7$\pm$1.7 & \textbf{77.5$\pm$1.5}\\
%		FMD & \textbf{77.3$\pm$1.4} & 77.3$\pm$1.6\\
%		UIUC & 97.5$\pm$0.7 & \textbf{98.0$\pm$0.8}\\
%		UMD & 98.1$\pm$0.9 & \textbf{98.4$\pm$0.8}\\		
%		1200Tex & \textbf{87.8$\pm$1.3} & 87.3$\pm$1.2\\
		\hline		
	\end{tabular}
\end{table} 

\begin{figure}[!htpb]
	\begin{tabular}{cc}
		\includegraphics[width=.45\textwidth]{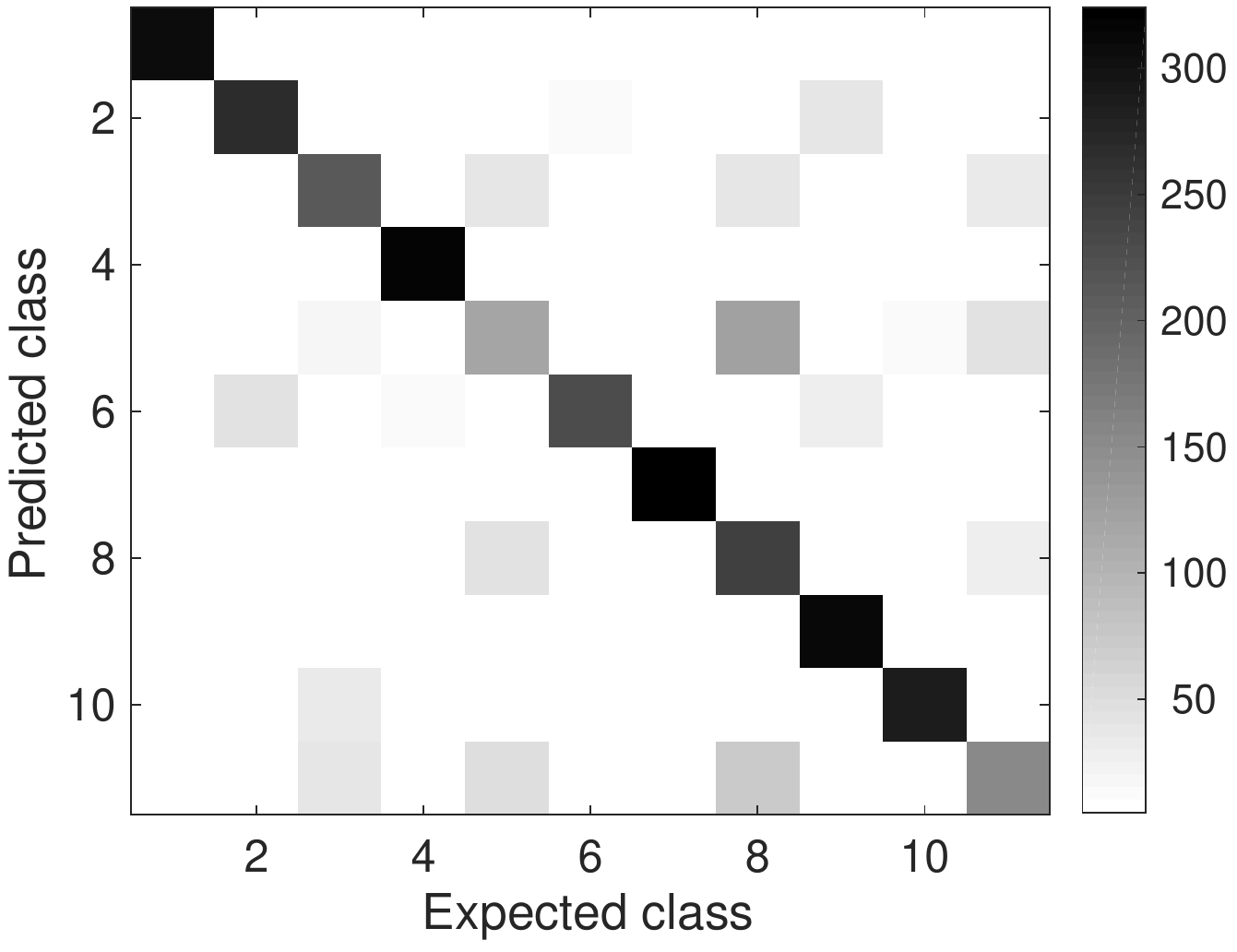} &
		\includegraphics[width=.45\textwidth]{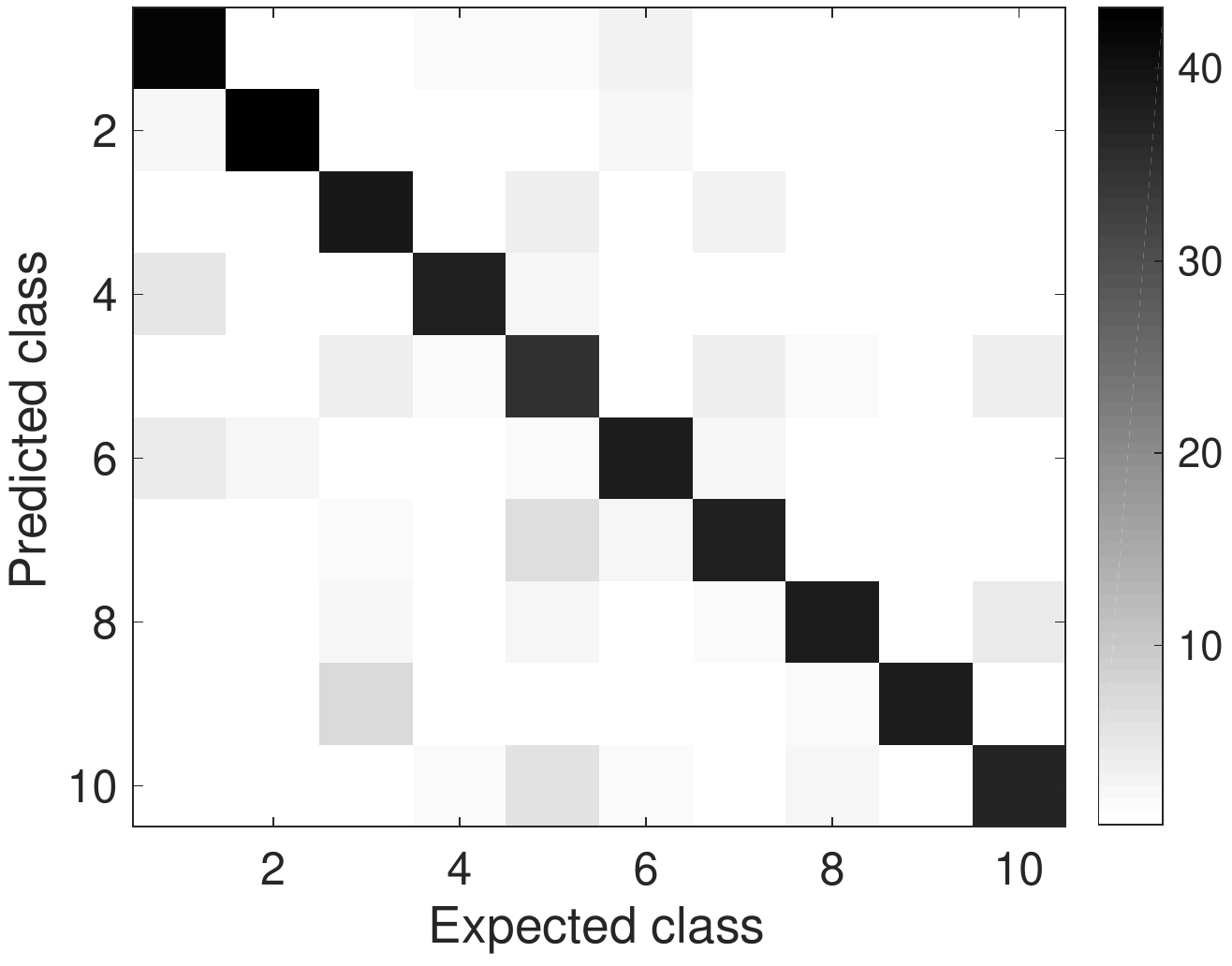}\\
		(a) & (b)\\
		\includegraphics[width=.45\textwidth]{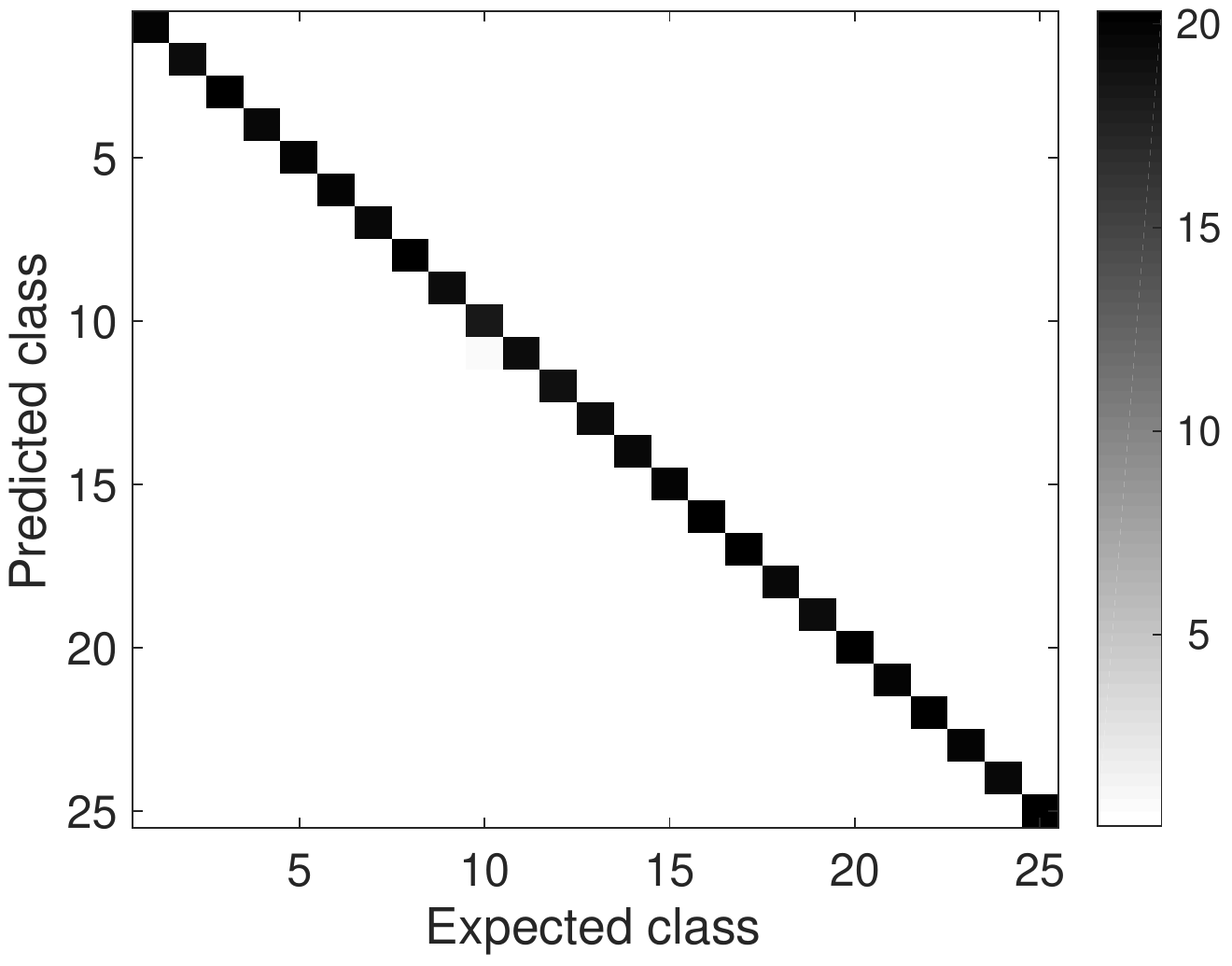} &
		\includegraphics[width=.45\textwidth]{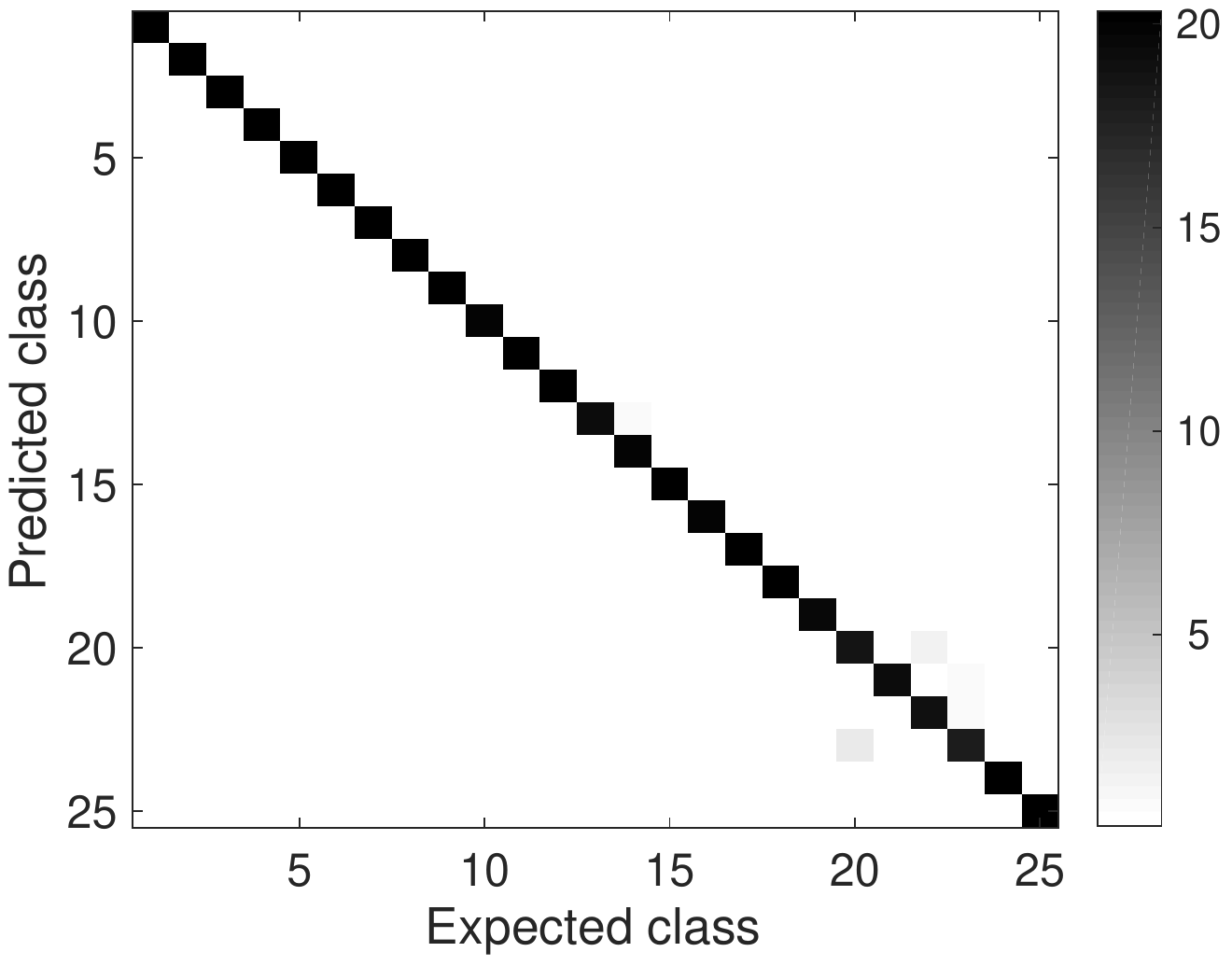}\\
		(c) & (d)\\
	\end{tabular}
	\caption{Confusion matrices for the proposed VisGraphNet descriptors (H1+DS). (a) KTHTIPS-2b. (b) FMD. (c) UIUC. (d) UMD.}
	\label{fig:CM1}
\end{figure}

Figure \ref{fig:CM1} shows the confusion matrices for the proposed method on the benchmark data sets. The combination H1+DS was chosen for this analysis as according to the previous tests it presented the best performance in most cases. Figure \ref{fig:CM1} (c) and (d) correspond to a nearly ideal classification result, confirming the excellent performance on UIUC and UMD attested in Table \ref{tab:SRdatabase}. Figure \ref{fig:CM1} (a) and (b), on the other hand, express a much more challenging scenario, where the number of misclassified samples (gray squares outside the diagonal) is significantly larger. This also confirms the lower accuracy achieved for KTHTIPS-2b and FMD. In FMD the misclassification occurrences are more evenly distributed along the different categories of materials, whereas in KTHTIPS-2b we have materials sensibly more challenging than others. Examples are classes 5 and 10 (respectively, ``cotton'' and ``wool''), which are often confused. Those materials share remarkable similarities especially in their microtextures. Both contain periodic patterns that are hardly distinguished sometimes even by visual inspection.

Table \ref{tab:SRdatabase} lists the accuracy performance of the VisGraphNet descriptors (H1 + DS) in KTHTIPS-2b, FMD, UIUC, and UMD, compared with other results published in the literature. In general, the proposed method demonstrates competitiveness with the state-of-the art, presenting overall better performance than the other ones in this comparison. Even in cases when VisGraphNet is outperformed in UIUC and UMD, it is by a marginal difference and the competitor approach involves complex combinations at the level of CNN architecture, which are pretty expensive in computational terms. In particular, the noticeable accuracy in KTHTIPS-2b and FMD is particularly relevant as those are databases well known for the challenge imposed to any recognition algorithm.
\begin{table}[!htpb]
	\centering
	\caption{Accuracy of the proposed descriptors (H1+DS) compared with other texture descriptors in the literature. All the results except for the proposed VisGraphNet were obtained from the literature. A `-' indicates that no result was published for that method on that database. A superscript $^1$ in KTH-TIPS2b represents a slightly different training/testing split where 3 samples are used for training and the remaining one for testing.}
	\label{tab:SRdatabase}
	\scalebox{0.9}{
	\begin{tabular}{ccccc}
		Method & KTH-TIPS2b & FMD & UIUC & UMD\\
		\hline
		VZ-MR8 \cite{VZ05} & 46.3 & 22.1 & 92.9 & - \\
		LBP \cite{OPM02} & 50.5 & - & 88.4 & 96.1\\
		VZ-Joint \cite{VZ09} & 53.3 & 23.8 & 78.4 & - \\
		LBP-FH \cite{AMHP09} & 54.6 & - & - & - \\
		CLBP \cite{GZZ10} & 57.3 & 43.6 & 95.7 & 98.6 \\
		SIFT+LLC \cite{CMKV16} & 57.6 & 50.4 & 96.3 & 98.4\\					
		ELBP \cite{LZLKF12} & 58.1 & 58.1 & - & - \\
		SIFT + KCB \cite{CMKMV14} & 58.3 & 45.1 & 91.4 & 98.0\\
		SIFT + BoVW \cite{CMKMV14} & 58.4 & 49.5 & 96.1 & 98.1\\
		SIFT + VLAD \cite{CMKMV14} & 63.1 & 52.6 & 96.5 & 99.3\\
		RandNet (NNC) \cite{CJGLZM15} & 60.7$^1$ & - & 56.6 & 90.9\\
		PCANet (NNC) \cite{CJGLZM15} & 59.4$^1$ & - & 57.7 & 90.5\\
		BSIF \cite{KR12} & 54.3 & - & 73.4 & 96.1\\			
		LBP$_{riu2}$/VAR \cite{OPM02} & 58.5$^1$ & - & 84.4 & 95.9\\
		SIFT+IFV \cite{CMKMV14} & 58.2 & 69.3 & 97.0 & 99.2 \\
		ScatNet (NNC) \cite{BM13} & 63.7$^1$ & - & 88.6 & 93.4\\
		FC-CNN AlexNet \cite{CMKV16} & 71.5 & 64.8 & 91.1 & 95.9\\
		MFS \cite{XJF09} & - & - & 92.7 & 93.9 \\
		DeCAF \cite{CMKMV14} & 70.7 & 70.7 & 94.2 & 96.4\\
		FC-CNN VGGM \cite{CMKV16} & 71.0 & 70.3 & 94.5 & 97.2\\			
		(H+L)(S+R) \cite{LSP05} & - & - & 97.0 & 97.0\\			
		FC-CNN + FV-CNN AlexNet \cite{CMKV16} & 72.1 & 71.4 & 99.3 & 99.7\\
		FC-CNN VGGVD \cite{CMKV16} & 75.4 & 75.0 & 99.9 & 97.7\\			
		\hline
		VisGraphNet (Proposed) & 77.5 & 77.3 & 98.0 & 98.4\\
		\hline		
	\end{tabular}
	}
\end{table}

\subsection{Identification of Plant Species}

Table \ref{tab:SRdatabase_plant} shows the performance of the proposed descriptors in 1200Tex database \cite{CMB09}, compared with other state-of-the-art results in the literature on these textures. This is a set of images of plant leaves from 20 Brazilian species collected \textit{in vivo}. For each species 20 samples were collected, cleaned, aligned with the vertical axis and photographed by a commercial scanner. The image of each sample was divided into 3 non-overlapping windows with size $128\times 128$. Those windows were extracted from regions of the leaf presenting less texture variance and were converted into gray scale images, resulting in a total of 1200 images equally distributed among 20 plant species. The training/testing split was similar to that used for UMD and UIUC, i.e., for each species 30 images were randomly selected for training and the remaining images for testing. Such random division was repeated 10 times to provide average accuracy and deviation. 

Some performance gain is observed over the original FC-CNN VGGVD descriptors when the visibility graph modeling is employed. While the classical deep learning approach achieved an accuracy of 84.2\%, the VG model (H1+DS) provided 87.3\%.
\begin{table}[!htpb]
	\centering
	\caption{Accuracy of the VisGraphNet descriptors over 1200Tex database, compared with other results in the literature.}
	\label{tab:SRdatabase_plant}	
	\begin{tabular}{cc}
		\hline
		Method & Accuracy (\%)\\
		\hline
		LBPV \cite{GZZ10} & 70.8\\
		Network diffusion \cite{GSFB16} & 75.8\\
		FC-CNN VGGM \cite{CMKV16} & 78.0\\		
		Gabor \cite{GSFB16} & 84.0\\
		FC-CNN VGGVD \cite{CMKV16} & 84.2\\
		Schroedinger \cite{FB17} & 85.3\\		
		SIFT + BoVW \cite{CMKMV14} & 86.0\\		
		FV-CNN VGGVD \cite{CMKV16} & 87.1\\		
		\hline
		Proposed & 87.3\\
		\hline
	\end{tabular}
\end{table}	

\begin{figure}[!htpb]
	\centering
	\includegraphics[width=.45\textwidth]{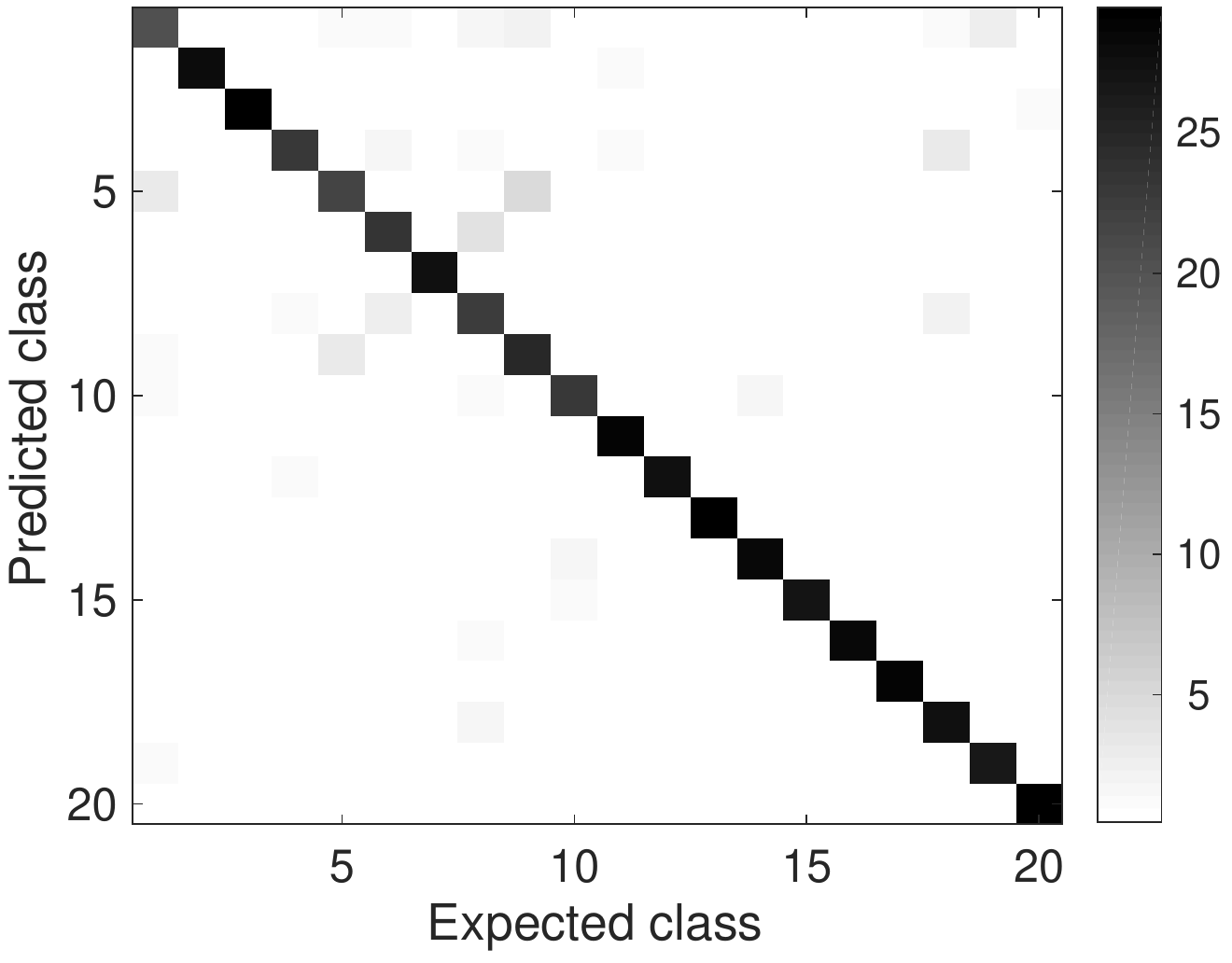}
	\caption{Confusion matrix for the plant database.}
	\label{fig:CM2}
\end{figure}

Figure \ref{fig:CM2} gives more information about the classification outcomes by means of the correspondent confusion matrix. Generally speaking, such map illustrates the high accuracy obtained in this task. In most classes the ratio of species correctly identified is close to 100\%. The most complicated cases in the matrix are between the pairs of classes 6/8 and 5/9. The respective leaves are characterized by similar patterns in nervure distribution, which are known to be fundamental source of information for species discrimination, such that lower accuracy were already expected in such situation.

Both in the benchmark and in the practical experiment we observed that VG modeling contributed with the classification process, improving the accuracy achieved by the original features (here, the CNN descriptors). Such gain was theoretically expected considering the ability of visibility graphs in identifying patterns of randomness, periodicity, fractality, chaoticity, etc. These are attributes whose importance in recognition tasks are known for a long time in the literature. The results confirm such positive expectations and points in the direction of more in-depth investigation on how tools from complex networks combined with neural networks could be useful for feature representation and classification in general.

\section{Conclusions}

We proposed and investigated a combination of feature map codes provided by a neural network as in \cite{CMKV16} with the modeling by visibility graphs (VG).

VGs are capable of providing a new perspective over the original data, in this way hopefully enriching the analysis and impacting the accuracy of the classification process.

The performance of our method was evaluated in benchmark databases as well as in a practical problem with biological importance, namely, the identification of species of Brazilian plants. In both situations the proposed method achieved promising results, being competitive when compared to state-of-the-art techniques recently published.

Such results suggest more in-depth investigation on how techniques from different contexts in data analysis (complex networks in our case) can be useful to explore the information conveyed by the highly non-linear features provided by a multilayer neural network. Our results confirm that the classification performance can be significantly improved in this way, especially in those more challenging applications.

\section*{Acknowledgements}

J. B. F. gratefully acknowledges the financial support of S\~ao Paulo Research Foundation (FAPESP) (Grant \#2016/16060-0) and from National Council for Scientific and Technological Development, Brazil (CNPq) (Grants \#301480/2016-8 and \#423292/2018-8).
This study was financed in part by the Coordena\c{c}\~{a}o de Aperfei\c{c}oamento de Pessoal de N\'{i}vel Superior - Brasil (CAPES) - Finance Code 001* - under the ``CAPES PrInt program''.

\section*{References}

%\bibliographystyle{model2-names}
%\bibliography{CNNVisibilityGraph}

\end{document}